\documentclass[twoside,11pt,fleqn]{article}
\usepackage{jair}
\usepackage{theapa}
\usepackage{times}
\usepackage{latexsym}
\usepackage{amsmath}
\usepackage{amsfonts}
\usepackage{theorem}
\usepackage{ifthen}
\usepackage{url}
\usepackage{nicefrac}

\input{pygment-settings}

\newcommand{\pre}{\operatorname{pre}}
\newcommand{\add}{\operatorname{add}}
\newcommand{\del}{\operatorname{del}}

\newcommand{\mptype}[1]{\text{{\small\textsf{#1}}}}
\newcommand{\ptype}[1]{{\small\textsf{#1}}}

\theorembodyfont{\rm}

\newcommand{\citeeg}[1]{\def\leftcite{(e.g.\/\ }\cite{#1}\def\leftcite{(}}

\begin{document}

\title{The Universal PDDL Domain}
\author{Patrik Haslum\\
  Australian National University \\
  {\tt firstname.lastname@anu.edu.au}\medskip\\
  Augusto B. Corr\^{e}a\\
  University of Basel, Switzerland \\
  {\tt augusto.blaascorrea@unibas.ch}}
\date{}
\maketitle

\begin{abstract}
In AI planning, it is common to distinguish between planning domains and problem
instances, where a ``domain'' is generally understood as a set of related
problem instances. This distinction is important, for example, in generalised
planning, which aims to find a single, general plan or policy that solves all
instances of a given domain. In PDDL, domains and problem instances are clearly
separated: the domain defines the types, predicate symbols, and action schemata,
while the problem instance specifies the concrete set of (typed) objects, the
initial state, and the goal condition. In this paper, we show that it is quite
easy to define a PDDL domain such that any propositional planning problem
instance, from any domain, becomes an instance of this (lifted) ``universal''
domain. We construct different formulations of the universal domain, and discuss
their implications for the complexity of lifted domain-dependent or generalised
planning.
\end{abstract}

\section{Introduction}

In AI planning, a distinction is often made between planning domains and
problem instances, where a ``domain'' is intuitively understood to be a
set, typically infinite, of related or similar problem instances.
This concept is important in, for instance, planning with domain-specific
control knowledge
\citeeg{bacchus-kabanza95,doherty-kvarnstrom99,wilkins-desjardins00},
and in generalised planning, which seeks a single, general plan or policy
that solves all instances of a given domain
\citeeg{srivastava-immerman-zilberstein:AIJ:2011}.
It is materialised in many modelling languages for specifying planning
problems, such as PDDL \cite{haslum-etal:pddl19}, in which the domain
and problem instance are syntactically separate.
In PDDL, the domain definition contains types, and parameterised predicates
and action schemata. The problem instance definition provides the concrete
set of (typed) objects, the initial state and the goal condition.

\citeA{grundke-roeger-helmert:ICAPS:2024} and \citeA{haslum-scholz:icaps03ws}
argue that PDDL's notion of domain is too weak, in that it does not always
allow the modeller to explicitly state the constraints necessary to define
precisely the intended set of problem instances, such as constraints on
intended ``valid'' initial states and goals.

Here, we will show that PDDL's notion of domain is also in another sense
too general: specifically, that it is possible (indeed, quite trivial) to
define a domain such that \emph{any} planning problem instance, of any
domain, is an instance of this ``universal'' domain.
There is, however, is caveat: While the universal domain is a parameterised
PDDL domain, consisting of types, predicates and action schemata, instances
of this domain are arbitrary propositional planning problems.
This means that although any PDDL domain--problem pair can be turned into
an instance of the universal domain, doing so requires grounding it, with
the consequent potentially exponential increase in size.
We will also argue that it is not possible to define a universal domain
in PDDL such that any domain--problem pair can be expressed as an instance
of this domain of a size that is polynomial in that of the domain--problem
pair.

\begin{figure}
\begin{Verbatim}[commandchars=\\\{\}]
\PY{p}{(}\PY{k}{define} \PY{p}{(}\PY{k}{domain} \PY{n+nx}{planning}\PY{p}{)}
  \PY{p}{(}\PY{k}{:types} \PY{n+nc}{action} \PY{n+nc}{proposition}\PY{p}{)}
  \PY{p}{(}\PY{k}{:predicates} \PY{p}{(}\PY{n+nx}{pre} \PY{n+nv}{?a} \PY{n+nc}{\PYZhy{}} \PY{n+nc}{action} \PY{n+nv}{?p} \PY{n+nc}{\PYZhy{}} \PY{n+nc}{proposition}\PY{p}{)}
	       \PY{p}{(}\PY{n+nx}{add} \PY{n+nv}{?a} \PY{n+nc}{\PYZhy{}} \PY{n+nc}{action} \PY{n+nv}{?p} \PY{n+nc}{\PYZhy{}} \PY{n+nc}{proposition}\PY{p}{)}
	       \PY{p}{(}\PY{n+nx}{del} \PY{n+nv}{?a} \PY{n+nc}{\PYZhy{}} \PY{n+nc}{action} \PY{n+nv}{?p} \PY{n+nc}{\PYZhy{}} \PY{n+nc}{proposition}\PY{p}{)}
	       \PY{p}{(}\PY{n+nx}{true} \PY{n+nv}{?p} \PY{n+nc}{\PYZhy{}} \PY{n+nc}{proposition}\PY{p}{)}\PY{p}{)}

  \PY{p}{(}\PY{k}{:action} \PY{n+nx}{apply}
   \PY{k}{:parameters} \PY{p}{(}\PY{n+nv}{?a} \PY{n+nc}{\PYZhy{}} \PY{n+nc}{action}\PY{p}{)}
   \PY{k}{:precondition} \PY{p}{(}\PY{n+nb}{forall} \PY{p}{(}\PY{n+nv}{?p} \PY{n+nc}{\PYZhy{}} \PY{n+nc}{proposition}\PY{p}{)}
			 \PY{p}{(}\PY{n+nb}{imply} \PY{p}{(}\PY{n+nx}{pre} \PY{n+nv}{?a} \PY{n+nv}{?p}\PY{p}{)} \PY{p}{(}\PY{n+nx}{true} \PY{n+nv}{?p}\PY{p}{)}\PY{p}{)}\PY{p}{)}
   \PY{k}{:effect} \PY{p}{(}\PY{n+nb}{and} \PY{p}{(}\PY{n+nb}{forall} \PY{p}{(}\PY{n+nv}{?p} \PY{n+nc}{\PYZhy{}} \PY{n+nc}{proposition}\PY{p}{)}
			\PY{p}{(}\PY{n+nb}{when} \PY{p}{(}\PY{n+nx}{add} \PY{n+nv}{?a} \PY{n+nv}{?p}\PY{p}{)}
			  \PY{p}{(}\PY{n+nx}{true} \PY{n+nv}{?p}\PY{p}{)}\PY{p}{)}\PY{p}{)}
		\PY{p}{(}\PY{n+nb}{forall} \PY{p}{(}\PY{n+nv}{?p} \PY{n+nc}{\PYZhy{}} \PY{n+nc}{proposition}\PY{p}{)}
			\PY{p}{(}\PY{n+nb}{when} \PY{p}{(}\PY{n+nb}{and} \PY{p}{(}\PY{n+nx}{del} \PY{n+nv}{?a} \PY{n+nv}{?p}\PY{p}{)} \PY{p}{(}\PY{n+nb}{not} \PY{p}{(}\PY{n+nx}{add} \PY{n+nv}{?a} \PY{n+nv}{?p}\PY{p}{)}\PY{p}{)}\PY{p}{)}
			  \PY{p}{(}\PY{n+nb}{not} \PY{p}{(}\PY{n+nx}{true} \PY{n+nv}{?p}\PY{p}{)}\PY{p}{)}\PY{p}{)}\PY{p}{)}\PY{p}{)}
   \PY{p}{)}
  \PY{p}{)}
\end{Verbatim}
\caption{The universal domain for propositional planning.}
\label{fig:upp}
\end{figure}

\begin{figure}
\tiny
\begin{Verbatim}[commandchars=\\\{\}]
\PY{p}{(}\PY{k}{define} \PY{p}{(}\PY{k}{problem} \PY{n+nx}{sussman}\PY{p}{)}
  \PY{p}{(}\PY{k}{:domain} \PY{n+nx}{planning}\PY{p}{)}
  \PY{p}{(}\PY{k}{:objects} \PY{n+nx}{ontable\PYZus{}A} \PY{n+nx}{ontable\PYZus{}B} \PY{n+nx}{ontable\PYZus{}C} \PY{n+nx}{on\PYZus{}A\PYZus{}B} \PY{n+nx}{on\PYZus{}A\PYZus{}C}
	    \PY{n+nx}{on\PYZus{}B\PYZus{}A} \PY{n+nx}{on\PYZus{}B\PYZus{}C} \PY{n+nx}{on\PYZus{}C\PYZus{}A} \PY{n+nx}{on\PYZus{}C\PYZus{}B} \PY{n+nx}{clear\PYZus{}A} \PY{n+nx}{clear\PYZus{}B} \PY{n+nx}{clear\PYZus{}C}
	    \PY{n+nx}{holding\PYZus{}A} \PY{n+nx}{holding\PYZus{}B} \PY{n+nx}{holding\PYZus{}C} \PY{n+nx}{hand\PYZus{}empty} \PY{n+nc}{\PYZhy{}} \PY{n+nc}{proposition}
	    \PY{n+nx}{pickup\PYZus{}A} \PY{n+nx}{pickup\PYZus{}B} \PY{n+nx}{pickup\PYZus{}C} \PY{n+nx}{putdown\PYZus{}A} \PY{n+nx}{putdown\PYZus{}B} \PY{n+nx}{putdown\PYZus{}C}
	    \PY{n+nx}{stack\PYZus{}A\PYZus{}B} \PY{n+nx}{stack\PYZus{}A\PYZus{}C} \PY{n+nx}{stack\PYZus{}B\PYZus{}A} \PY{n+nx}{stack\PYZus{}B\PYZus{}C} \PY{n+nx}{stack\PYZus{}C\PYZus{}A} \PY{n+nx}{stack\PYZus{}C\PYZus{}B}
	    \PY{n+nx}{unstack\PYZus{}A\PYZus{}B} \PY{n+nx}{unstack\PYZus{}A\PYZus{}C} \PY{n+nx}{unstack\PYZus{}B\PYZus{}A} \PY{n+nx}{unstack\PYZus{}B\PYZus{}C} \PY{n+nx}{unstack\PYZus{}C\PYZus{}A}
	    \PY{n+nx}{unstack\PYZus{}C\PYZus{}B} \PY{n+nc}{\PYZhy{}} \PY{n+nc}{action}\PY{p}{)}

  \PY{p}{(}\PY{k}{:init}
   \PY{p}{(}\PY{n+nx}{pre} \PY{n+nx}{pickup\PYZus{}A} \PY{n+nx}{ontable\PYZus{}A}\PY{p}{)} \PY{p}{(}\PY{n+nx}{pre} \PY{n+nx}{pickup\PYZus{}A} \PY{n+nx}{clear\PYZus{}A}\PY{p}{)} \PY{p}{(}\PY{n+nx}{pre} \PY{n+nx}{pickup\PYZus{}A} \PY{n+nx}{hand\PYZus{}emtpy}\PY{p}{)} \PY{p}{(}\PY{n+nx}{add} \PY{n+nx}{pickup\PYZus{}A} \PY{n+nx}{holding\PYZus{}A}\PY{p}{)}
   \PY{p}{(}\PY{n+nx}{del} \PY{n+nx}{pickup\PYZus{}A} \PY{n+nx}{ontable\PYZus{}A}\PY{p}{)} \PY{p}{(}\PY{n+nx}{del} \PY{n+nx}{pickup\PYZus{}A} \PY{n+nx}{clear\PYZus{}A}\PY{p}{)} \PY{p}{(}\PY{n+nx}{del} \PY{n+nx}{pickup\PYZus{}A} \PY{n+nx}{hand\PYZus{}empty}\PY{p}{)}
   \PY{p}{(}\PY{n+nx}{pre} \PY{n+nx}{pickup\PYZus{}B} \PY{n+nx}{ontable\PYZus{}B}\PY{p}{)} \PY{p}{(}\PY{n+nx}{pre} \PY{n+nx}{pickup\PYZus{}B} \PY{n+nx}{clear\PYZus{}B}\PY{p}{)} \PY{p}{(}\PY{n+nx}{pre} \PY{n+nx}{pickup\PYZus{}B} \PY{n+nx}{hand\PYZus{}emtpy}\PY{p}{)} \PY{p}{(}\PY{n+nx}{add} \PY{n+nx}{pickup\PYZus{}B} \PY{n+nx}{holding\PYZus{}B}\PY{p}{)}
   \PY{p}{(}\PY{n+nx}{del} \PY{n+nx}{pickup\PYZus{}B} \PY{n+nx}{ontable\PYZus{}B}\PY{p}{)} \PY{p}{(}\PY{n+nx}{del} \PY{n+nx}{pickup\PYZus{}B} \PY{n+nx}{clear\PYZus{}B}\PY{p}{)} \PY{p}{(}\PY{n+nx}{del} \PY{n+nx}{pickup\PYZus{}B} \PY{n+nx}{hand\PYZus{}empty}\PY{p}{)}
   \PY{p}{(}\PY{n+nx}{pre} \PY{n+nx}{pickup\PYZus{}C} \PY{n+nx}{ontable\PYZus{}C}\PY{p}{)} \PY{p}{(}\PY{n+nx}{pre} \PY{n+nx}{pickup\PYZus{}C} \PY{n+nx}{clear\PYZus{}C}\PY{p}{)} \PY{p}{(}\PY{n+nx}{pre} \PY{n+nx}{pickup\PYZus{}C} \PY{n+nx}{hand\PYZus{}emtpy}\PY{p}{)} \PY{p}{(}\PY{n+nx}{add} \PY{n+nx}{pickup\PYZus{}C} \PY{n+nx}{holding\PYZus{}C}\PY{p}{)}
   \PY{p}{(}\PY{n+nx}{del} \PY{n+nx}{pickup\PYZus{}C} \PY{n+nx}{ontable\PYZus{}C}\PY{p}{)} \PY{p}{(}\PY{n+nx}{del} \PY{n+nx}{pickup\PYZus{}C} \PY{n+nx}{clear\PYZus{}C}\PY{p}{)} \PY{p}{(}\PY{n+nx}{del} \PY{n+nx}{pickup\PYZus{}C} \PY{n+nx}{hand\PYZus{}empty}\PY{p}{)}
   \PY{p}{(}\PY{n+nx}{pre} \PY{n+nx}{putdown\PYZus{}A} \PY{n+nx}{holding\PYZus{}A}\PY{p}{)} \PY{p}{(}\PY{n+nx}{add} \PY{n+nx}{putdown\PYZus{}A} \PY{n+nx}{ontable\PYZus{}A}\PY{p}{)} \PY{p}{(}\PY{n+nx}{add} \PY{n+nx}{putown\PYZus{}A} \PY{n+nx}{clear\PYZus{}A}\PY{p}{)}
   \PY{p}{(}\PY{n+nx}{add} \PY{n+nx}{putdown\PYZus{}A} \PY{n+nx}{hand\PYZus{}empty}\PY{p}{)} \PY{p}{(}\PY{n+nx}{del} \PY{n+nx}{putdown\PYZus{}A} \PY{n+nx}{holding\PYZus{}A}\PY{p}{)}
   \PY{p}{(}\PY{n+nx}{pre} \PY{n+nx}{putdown\PYZus{}B} \PY{n+nx}{holding\PYZus{}B}\PY{p}{)} \PY{p}{(}\PY{n+nx}{add} \PY{n+nx}{putdown\PYZus{}B} \PY{n+nx}{ontable\PYZus{}B}\PY{p}{)} \PY{p}{(}\PY{n+nx}{add} \PY{n+nx}{putown\PYZus{}B} \PY{n+nx}{clear\PYZus{}B}\PY{p}{)}
   \PY{p}{(}\PY{n+nx}{add} \PY{n+nx}{putdown\PYZus{}B} \PY{n+nx}{hand\PYZus{}empty}\PY{p}{)} \PY{p}{(}\PY{n+nx}{del} \PY{n+nx}{putdown\PYZus{}B} \PY{n+nx}{holding\PYZus{}B}\PY{p}{)}
   \PY{p}{(}\PY{n+nx}{pre} \PY{n+nx}{putdown\PYZus{}C} \PY{n+nx}{holding\PYZus{}C}\PY{p}{)} \PY{p}{(}\PY{n+nx}{add} \PY{n+nx}{putdown\PYZus{}C} \PY{n+nx}{ontable\PYZus{}C}\PY{p}{)} \PY{p}{(}\PY{n+nx}{add} \PY{n+nx}{putown\PYZus{}C} \PY{n+nx}{clear\PYZus{}C}\PY{p}{)}
   \PY{p}{(}\PY{n+nx}{add} \PY{n+nx}{putdown\PYZus{}C} \PY{n+nx}{hand\PYZus{}empty}\PY{p}{)} \PY{p}{(}\PY{n+nx}{del} \PY{n+nx}{putdown\PYZus{}C} \PY{n+nx}{holding\PYZus{}C}\PY{p}{)}
   \PY{p}{(}\PY{n+nx}{pre} \PY{n+nx}{stack\PYZus{}A\PYZus{}B} \PY{n+nx}{holding\PYZus{}A}\PY{p}{)} \PY{p}{(}\PY{n+nx}{pre} \PY{n+nx}{stack\PYZus{}A\PYZus{}B} \PY{n+nx}{clear\PYZus{}B}\PY{p}{)} \PY{p}{(}\PY{n+nx}{add} \PY{n+nx}{stack\PYZus{}A\PYZus{}B} \PY{n+nx}{on\PYZus{}A\PYZus{}B}\PY{p}{)} \PY{p}{(}\PY{n+nx}{add} \PY{n+nx}{stack\PYZus{}A\PYZus{}B} \PY{n+nx}{clear\PYZus{}A}\PY{p}{)}
   \PY{p}{(}\PY{n+nx}{add} \PY{n+nx}{stack\PYZus{}A\PYZus{}B} \PY{n+nx}{hand\PYZus{}empty}\PY{p}{)} \PY{p}{(}\PY{n+nx}{del} \PY{n+nx}{stack\PYZus{}A\PYZus{}B} \PY{n+nx}{holding\PYZus{}A}\PY{p}{)} \PY{p}{(}\PY{n+nx}{del} \PY{n+nx}{stack\PYZus{}A\PYZus{}B} \PY{n+nx}{clear\PYZus{}B}\PY{p}{)}
   \PY{p}{(}\PY{n+nx}{pre} \PY{n+nx}{stack\PYZus{}A\PYZus{}C} \PY{n+nx}{holding\PYZus{}A}\PY{p}{)} \PY{p}{(}\PY{n+nx}{pre} \PY{n+nx}{stack\PYZus{}A\PYZus{}C} \PY{n+nx}{clear\PYZus{}C}\PY{p}{)} \PY{p}{(}\PY{n+nx}{add} \PY{n+nx}{stack\PYZus{}A\PYZus{}C} \PY{n+nx}{on\PYZus{}A\PYZus{}C}\PY{p}{)} \PY{p}{(}\PY{n+nx}{add} \PY{n+nx}{stack\PYZus{}A\PYZus{}C} \PY{n+nx}{clear\PYZus{}A}\PY{p}{)}
   \PY{p}{(}\PY{n+nx}{add} \PY{n+nx}{stack\PYZus{}A\PYZus{}C} \PY{n+nx}{hand\PYZus{}empty}\PY{p}{)} \PY{p}{(}\PY{n+nx}{del} \PY{n+nx}{stack\PYZus{}A\PYZus{}C} \PY{n+nx}{holding\PYZus{}A}\PY{p}{)} \PY{p}{(}\PY{n+nx}{del} \PY{n+nx}{stack\PYZus{}A\PYZus{}C} \PY{n+nx}{clear\PYZus{}C}\PY{p}{)}
   \PY{p}{(}\PY{n+nx}{pre} \PY{n+nx}{stack\PYZus{}B\PYZus{}A} \PY{n+nx}{holding\PYZus{}B}\PY{p}{)} \PY{p}{(}\PY{n+nx}{pre} \PY{n+nx}{stack\PYZus{}B\PYZus{}A} \PY{n+nx}{clear\PYZus{}A}\PY{p}{)} \PY{p}{(}\PY{n+nx}{add} \PY{n+nx}{stack\PYZus{}B\PYZus{}A} \PY{n+nx}{on\PYZus{}B\PYZus{}A}\PY{p}{)} \PY{p}{(}\PY{n+nx}{add} \PY{n+nx}{stack\PYZus{}B\PYZus{}A} \PY{n+nx}{clear\PYZus{}B}\PY{p}{)}
   \PY{p}{(}\PY{n+nx}{add} \PY{n+nx}{stack\PYZus{}B\PYZus{}A} \PY{n+nx}{hand\PYZus{}empty}\PY{p}{)} \PY{p}{(}\PY{n+nx}{del} \PY{n+nx}{stack\PYZus{}B\PYZus{}A} \PY{n+nx}{holding\PYZus{}B}\PY{p}{)} \PY{p}{(}\PY{n+nx}{del} \PY{n+nx}{stack\PYZus{}B\PYZus{}A} \PY{n+nx}{clear\PYZus{}A}\PY{p}{)}
   \PY{p}{(}\PY{n+nx}{pre} \PY{n+nx}{stack\PYZus{}B\PYZus{}C} \PY{n+nx}{holding\PYZus{}B}\PY{p}{)} \PY{p}{(}\PY{n+nx}{pre} \PY{n+nx}{stack\PYZus{}B\PYZus{}C} \PY{n+nx}{clear\PYZus{}C}\PY{p}{)} \PY{p}{(}\PY{n+nx}{add} \PY{n+nx}{stack\PYZus{}B\PYZus{}C} \PY{n+nx}{on\PYZus{}B\PYZus{}C}\PY{p}{)} \PY{p}{(}\PY{n+nx}{add} \PY{n+nx}{stack\PYZus{}B\PYZus{}C} \PY{n+nx}{clear\PYZus{}B}\PY{p}{)}
   \PY{p}{(}\PY{n+nx}{add} \PY{n+nx}{stack\PYZus{}B\PYZus{}C} \PY{n+nx}{hand\PYZus{}empty}\PY{p}{)} \PY{p}{(}\PY{n+nx}{del} \PY{n+nx}{stack\PYZus{}B\PYZus{}C} \PY{n+nx}{holding\PYZus{}B}\PY{p}{)} \PY{p}{(}\PY{n+nx}{del} \PY{n+nx}{stack\PYZus{}B\PYZus{}C} \PY{n+nx}{clear\PYZus{}C}\PY{p}{)}
   \PY{p}{(}\PY{n+nx}{pre} \PY{n+nx}{stack\PYZus{}C\PYZus{}A} \PY{n+nx}{holding\PYZus{}C}\PY{p}{)} \PY{p}{(}\PY{n+nx}{pre} \PY{n+nx}{stack\PYZus{}C\PYZus{}A} \PY{n+nx}{clear\PYZus{}A}\PY{p}{)} \PY{p}{(}\PY{n+nx}{add} \PY{n+nx}{stack\PYZus{}C\PYZus{}A} \PY{n+nx}{on\PYZus{}C\PYZus{}A}\PY{p}{)} \PY{p}{(}\PY{n+nx}{add} \PY{n+nx}{stack\PYZus{}C\PYZus{}A} \PY{n+nx}{clear\PYZus{}C}\PY{p}{)}
   \PY{p}{(}\PY{n+nx}{add} \PY{n+nx}{stack\PYZus{}C\PYZus{}A} \PY{n+nx}{hand\PYZus{}empty}\PY{p}{)} \PY{p}{(}\PY{n+nx}{del} \PY{n+nx}{stack\PYZus{}C\PYZus{}A} \PY{n+nx}{holding\PYZus{}C}\PY{p}{)} \PY{p}{(}\PY{n+nx}{del} \PY{n+nx}{stack\PYZus{}C\PYZus{}A} \PY{n+nx}{clear\PYZus{}A}\PY{p}{)}
   \PY{p}{(}\PY{n+nx}{pre} \PY{n+nx}{stack\PYZus{}C\PYZus{}B} \PY{n+nx}{holding\PYZus{}C}\PY{p}{)} \PY{p}{(}\PY{n+nx}{pre} \PY{n+nx}{stack\PYZus{}C\PYZus{}B} \PY{n+nx}{clear\PYZus{}B}\PY{p}{)} \PY{p}{(}\PY{n+nx}{add} \PY{n+nx}{stack\PYZus{}C\PYZus{}B} \PY{n+nx}{on\PYZus{}C\PYZus{}B}\PY{p}{)} \PY{p}{(}\PY{n+nx}{add} \PY{n+nx}{stack\PYZus{}C\PYZus{}B} \PY{n+nx}{clear\PYZus{}C}\PY{p}{)}
   \PY{p}{(}\PY{n+nx}{add} \PY{n+nx}{stack\PYZus{}C\PYZus{}B} \PY{n+nx}{hand\PYZus{}empty}\PY{p}{)} \PY{p}{(}\PY{n+nx}{del} \PY{n+nx}{stack\PYZus{}C\PYZus{}B} \PY{n+nx}{holding\PYZus{}C}\PY{p}{)} \PY{p}{(}\PY{n+nx}{del} \PY{n+nx}{stack\PYZus{}C\PYZus{}B} \PY{n+nx}{clear\PYZus{}B}\PY{p}{)}
   \PY{p}{(}\PY{n+nx}{pre} \PY{n+nx}{unstack\PYZus{}A\PYZus{}B} \PY{n+nx}{on\PYZus{}A\PYZus{}B}\PY{p}{)} \PY{p}{(}\PY{n+nx}{pre} \PY{n+nx}{unstack\PYZus{}A\PYZus{}B} \PY{n+nx}{clear\PYZus{}A}\PY{p}{)} \PY{p}{(}\PY{n+nx}{pre} \PY{n+nx}{unstack\PYZus{}A\PYZus{}B} \PY{n+nx}{hand\PYZus{}empty}\PY{p}{)} \PY{p}{(}\PY{n+nx}{add} \PY{n+nx}{unstack\PYZus{}A\PYZus{}B} \PY{n+nx}{holding\PYZus{}A}\PY{p}{)}
   \PY{p}{(}\PY{n+nx}{add} \PY{n+nx}{unstack\PYZus{}A\PYZus{}B} \PY{n+nx}{clear\PYZus{}B}\PY{p}{)} \PY{p}{(}\PY{n+nx}{del} \PY{n+nx}{unstack\PYZus{}A\PYZus{}B} \PY{n+nx}{on\PYZus{}A\PYZus{}B}\PY{p}{)} \PY{p}{(}\PY{n+nx}{del} \PY{n+nx}{unstack\PYZus{}A\PYZus{}B} \PY{n+nx}{clear\PYZus{}A}\PY{p}{)}
   \PY{p}{(}\PY{n+nx}{pre} \PY{n+nx}{unstack\PYZus{}A\PYZus{}C} \PY{n+nx}{on\PYZus{}A\PYZus{}C}\PY{p}{)} \PY{p}{(}\PY{n+nx}{pre} \PY{n+nx}{unstack\PYZus{}A\PYZus{}C} \PY{n+nx}{clear\PYZus{}A}\PY{p}{)} \PY{p}{(}\PY{n+nx}{pre} \PY{n+nx}{unstack\PYZus{}A\PYZus{}C} \PY{n+nx}{hand\PYZus{}empty}\PY{p}{)} \PY{p}{(}\PY{n+nx}{add} \PY{n+nx}{unstack\PYZus{}A\PYZus{}C} \PY{n+nx}{holding\PYZus{}A}\PY{p}{)}
   \PY{p}{(}\PY{n+nx}{add} \PY{n+nx}{unstack\PYZus{}A\PYZus{}C} \PY{n+nx}{clear\PYZus{}C}\PY{p}{)} \PY{p}{(}\PY{n+nx}{del} \PY{n+nx}{unstack\PYZus{}A\PYZus{}C} \PY{n+nx}{on\PYZus{}A\PYZus{}C}\PY{p}{)} \PY{p}{(}\PY{n+nx}{del} \PY{n+nx}{unstack\PYZus{}A\PYZus{}C} \PY{n+nx}{clear\PYZus{}A}\PY{p}{)}
   \PY{p}{(}\PY{n+nx}{pre} \PY{n+nx}{unstack\PYZus{}B\PYZus{}A} \PY{n+nx}{on\PYZus{}B\PYZus{}A}\PY{p}{)} \PY{p}{(}\PY{n+nx}{pre} \PY{n+nx}{unstack\PYZus{}B\PYZus{}A} \PY{n+nx}{clear\PYZus{}B}\PY{p}{)} \PY{p}{(}\PY{n+nx}{pre} \PY{n+nx}{unstack\PYZus{}B\PYZus{}A} \PY{n+nx}{hand\PYZus{}empty}\PY{p}{)} \PY{p}{(}\PY{n+nx}{add} \PY{n+nx}{unstack\PYZus{}B\PYZus{}A} \PY{n+nx}{holding\PYZus{}B}\PY{p}{)}
   \PY{p}{(}\PY{n+nx}{add} \PY{n+nx}{unstack\PYZus{}B\PYZus{}A} \PY{n+nx}{clear\PYZus{}A}\PY{p}{)} \PY{p}{(}\PY{n+nx}{del} \PY{n+nx}{unstack\PYZus{}B\PYZus{}A} \PY{n+nx}{on\PYZus{}B\PYZus{}A}\PY{p}{)} \PY{p}{(}\PY{n+nx}{del} \PY{n+nx}{unstack\PYZus{}B\PYZus{}A} \PY{n+nx}{clear\PYZus{}B}\PY{p}{)}
   \PY{p}{(}\PY{n+nx}{pre} \PY{n+nx}{unstack\PYZus{}B\PYZus{}C} \PY{n+nx}{on\PYZus{}B\PYZus{}C}\PY{p}{)} \PY{p}{(}\PY{n+nx}{pre} \PY{n+nx}{unstack\PYZus{}B\PYZus{}C} \PY{n+nx}{clear\PYZus{}B}\PY{p}{)} \PY{p}{(}\PY{n+nx}{pre} \PY{n+nx}{unstack\PYZus{}B\PYZus{}C} \PY{n+nx}{hand\PYZus{}empty}\PY{p}{)} \PY{p}{(}\PY{n+nx}{add} \PY{n+nx}{unstack\PYZus{}B\PYZus{}C} \PY{n+nx}{holding\PYZus{}B}\PY{p}{)}
   \PY{p}{(}\PY{n+nx}{add} \PY{n+nx}{unstack\PYZus{}B\PYZus{}C} \PY{n+nx}{clear\PYZus{}C}\PY{p}{)} \PY{p}{(}\PY{n+nx}{del} \PY{n+nx}{unstack\PYZus{}B\PYZus{}C} \PY{n+nx}{on\PYZus{}B\PYZus{}C}\PY{p}{)} \PY{p}{(}\PY{n+nx}{del} \PY{n+nx}{unstack\PYZus{}B\PYZus{}C} \PY{n+nx}{clear\PYZus{}B}\PY{p}{)}
   \PY{p}{(}\PY{n+nx}{pre} \PY{n+nx}{unstack\PYZus{}C\PYZus{}A} \PY{n+nx}{on\PYZus{}C\PYZus{}A}\PY{p}{)} \PY{p}{(}\PY{n+nx}{pre} \PY{n+nx}{unstack\PYZus{}C\PYZus{}A} \PY{n+nx}{clear\PYZus{}C}\PY{p}{)} \PY{p}{(}\PY{n+nx}{pre} \PY{n+nx}{unstack\PYZus{}C\PYZus{}A} \PY{n+nx}{hand\PYZus{}empty}\PY{p}{)} \PY{p}{(}\PY{n+nx}{add} \PY{n+nx}{unstack\PYZus{}C\PYZus{}A} \PY{n+nx}{holding\PYZus{}C}\PY{p}{)}
   \PY{p}{(}\PY{n+nx}{add} \PY{n+nx}{unstack\PYZus{}C\PYZus{}A} \PY{n+nx}{clear\PYZus{}A}\PY{p}{)} \PY{p}{(}\PY{n+nx}{del} \PY{n+nx}{unstack\PYZus{}C\PYZus{}A} \PY{n+nx}{on\PYZus{}C\PYZus{}A}\PY{p}{)} \PY{p}{(}\PY{n+nx}{del} \PY{n+nx}{unstack\PYZus{}C\PYZus{}A} \PY{n+nx}{clear\PYZus{}C}\PY{p}{)}
   \PY{p}{(}\PY{n+nx}{pre} \PY{n+nx}{unstack\PYZus{}C\PYZus{}B} \PY{n+nx}{on\PYZus{}C\PYZus{}B}\PY{p}{)} \PY{p}{(}\PY{n+nx}{pre} \PY{n+nx}{unstack\PYZus{}C\PYZus{}B} \PY{n+nx}{clear\PYZus{}C}\PY{p}{)} \PY{p}{(}\PY{n+nx}{pre} \PY{n+nx}{unstack\PYZus{}C\PYZus{}B} \PY{n+nx}{hand\PYZus{}empty}\PY{p}{)} \PY{p}{(}\PY{n+nx}{add} \PY{n+nx}{unstack\PYZus{}C\PYZus{}B} \PY{n+nx}{holding\PYZus{}C}\PY{p}{)}
   \PY{p}{(}\PY{n+nx}{add} \PY{n+nx}{unstack\PYZus{}C\PYZus{}B} \PY{n+nx}{clear\PYZus{}B}\PY{p}{)} \PY{p}{(}\PY{n+nx}{del} \PY{n+nx}{unstack\PYZus{}C\PYZus{}B} \PY{n+nx}{on\PYZus{}C\PYZus{}B}\PY{p}{)} \PY{p}{(}\PY{n+nx}{del} \PY{n+nx}{unstack\PYZus{}C\PYZus{}B} \PY{n+nx}{clear\PYZus{}C}\PY{p}{)}

   \PY{p}{(}\PY{n+nx}{true} \PY{n+nx}{ontable\PYZus{}A}\PY{p}{)} \PY{p}{(}\PY{n+nx}{true} \PY{n+nx}{on\PYZus{}C\PYZus{}A}\PY{p}{)} \PY{p}{(}\PY{n+nx}{true} \PY{n+nx}{clear\PYZus{}C}\PY{p}{)} \PY{p}{(}\PY{n+nx}{true} \PY{n+nx}{ontable\PYZus{}B}\PY{p}{)} \PY{p}{(}\PY{n+nx}{true} \PY{n+nx}{clear\PYZus{}B}\PY{p}{)}\PY{p}{)}

  \PY{p}{(}\PY{k}{:goal} \PY{p}{(}\PY{n+nb}{and} \PY{p}{(}\PY{n+nx}{true} \PY{n+nx}{on\PYZus{}A\PYZus{}B}\PY{p}{)} \PY{p}{(}\PY{n+nx}{true} \PY{n+nx}{on\PYZus{}B\PYZus{}C}\PY{p}{)}\PY{p}{)}\PY{p}{)}
  \PY{p}{)}
\end{Verbatim}
\caption{Example of a problem instance of the universal propositional
  planning domain.}
\label{fig:ex1}
\end{figure}

\section{The Universal Propositional Planning Domain}

The universal PDDL domain for propositional planning, in its simplest
form, is shown in Figure \ref{fig:upp}. It has two types: \ptype{action}
and \ptype{proposition}, representing the (ground) actions and propositions
of the problem instance, respectively. It has a single action schema,
\ptype{apply}, with one parameter \ptype{?a} of type \ptype{action}.
An instance of this action schema with $\mptype{?a} = a$ is applicable
iff every proposition of the problem instance is either true in the
current state or not a precondition of the action $a$, i.e., iff $a$
is applicable. Its effect is to make true every proposition that is an
add effect of $a$, and false every proposition that is a delete effect,
and not an add effect, of $a$ (implementing PDDL's delete-before-add
semantics). Thus, the effect of $(\mptype{apply}\ a)$ is exactly the
effect of $a$.

Given a propositional planning problem instance $\Pi$, an instance
$P_{\Pi}$ of the universal domain is constructed with all ground actions
and propositions in $\Pi$ as objects, initial state facts
$(\mptype{pre}\ a\ p)$, $(\mptype{add}\ a\ p)$ and $(\mptype{del}\ a\ p)$
for all ground actions $a$ and propositions $p$ such that
$p \in \pre(a)$, $p \in \add(a)$ and $p \in \del(a)$, respectively,
and $(\mptype{true}\ p)$ for each proposition true in the initial
state of $\Pi$, and goal
$(\mptype{and}\ (\mptype{true}\ p_1) \ldots (\mptype{true}\ p_m)$,
where $p_1, \ldots, p_m$ are the goal facts of $\Pi$.
It is easy to see that this instance has a plan iff $\Pi$ has a plan,
and the plan for $\Pi$ is in fact simply the sequence of arguments of
the actions in the plan for $P_\Pi$.

An example of an instance of the universal domain, representing a
classical planning problem, is shown in Figure \ref{fig:ex1}.

\begin{figure}
\tiny
\begin{Verbatim}[commandchars=\\\{\}]
\PY{p}{(}\PY{k}{define} \PY{p}{(}\PY{k}{domain} \PY{n+nx}{planning}\PY{p}{)}
  \PY{p}{(}\PY{k}{:types} \PY{n+nc}{action} \PY{n+nc}{proposition}\PY{p}{)}
  \PY{p}{(}\PY{k}{:predictes} \PY{p}{(}\PY{n+nx}{first\PYZhy{}pre} \PY{n+nv}{?a} \PY{n+nc}{\PYZhy{}} \PY{n+nc}{action} \PY{n+nv}{?p} \PY{n+nc}{\PYZhy{}} \PY{n+nc}{proposition}\PY{p}{)}
	      \PY{p}{(}\PY{n+nx}{next\PYZhy{}pre} \PY{n+nv}{?a} \PY{n+nc}{\PYZhy{}} \PY{n+nc}{action} \PY{n+nv}{?p} \PY{n+nc}{\PYZhy{}} \PY{n+nc}{proposition} \PY{n+nv}{?q} \PY{n+nc}{\PYZhy{}} \PY{n+nc}{proposition}\PY{p}{)}
	      \PY{p}{(}\PY{n+nx}{last\PYZhy{}pre} \PY{n+nv}{?a} \PY{n+nc}{\PYZhy{}} \PY{n+nc}{action} \PY{n+nv}{?p} \PY{n+nc}{\PYZhy{}} \PY{n+nc}{proposition}\PY{p}{)}
	      \PY{p}{(}\PY{n+nx}{has\PYZhy{}no\PYZhy{}pre} \PY{n+nv}{?a}\PY{p}{)}
	      \PY{p}{(}\PY{n+nx}{first\PYZhy{}add} \PY{n+nv}{?a} \PY{n+nc}{\PYZhy{}} \PY{n+nc}{action} \PY{n+nv}{?p} \PY{n+nc}{\PYZhy{}} \PY{n+nc}{proposition}\PY{p}{)}
	      \PY{p}{(}\PY{n+nx}{next\PYZhy{}add} \PY{n+nv}{?a} \PY{n+nc}{\PYZhy{}} \PY{n+nc}{action} \PY{n+nv}{?p} \PY{n+nc}{\PYZhy{}} \PY{n+nc}{proposition} \PY{n+nv}{?q} \PY{n+nc}{\PYZhy{}} \PY{n+nc}{proposition}\PY{p}{)}
	      \PY{p}{(}\PY{n+nx}{last\PYZhy{}add} \PY{n+nv}{?a} \PY{n+nc}{\PYZhy{}} \PY{n+nc}{action} \PY{n+nv}{?p} \PY{n+nc}{\PYZhy{}} \PY{n+nc}{proposition}\PY{p}{)}
	      \PY{p}{(}\PY{n+nx}{first\PYZhy{}del} \PY{n+nv}{?a} \PY{n+nc}{\PYZhy{}} \PY{n+nc}{action} \PY{n+nv}{?p} \PY{n+nc}{\PYZhy{}} \PY{n+nc}{proposition}\PY{p}{)}
	      \PY{p}{(}\PY{n+nx}{next\PYZhy{}del} \PY{n+nv}{?a} \PY{n+nc}{\PYZhy{}} \PY{n+nc}{action} \PY{n+nv}{?p} \PY{n+nc}{\PYZhy{}} \PY{n+nc}{proposition} \PY{n+nv}{?q} \PY{n+nc}{\PYZhy{}} \PY{n+nc}{proposition}\PY{p}{)}
	      \PY{p}{(}\PY{n+nx}{last\PYZhy{}del} \PY{n+nv}{?a} \PY{n+nc}{\PYZhy{}} \PY{n+nc}{action} \PY{n+nv}{?p} \PY{n+nc}{\PYZhy{}} \PY{n+nc}{proposition}\PY{p}{)}
	      \PY{p}{(}\PY{n+nx}{has\PYZhy{}no\PYZhy{}del} \PY{n+nv}{?a}\PY{p}{)}
	      \PY{p}{(}\PY{n+nx}{true} \PY{n+nv}{?p} \PY{n+nc}{\PYZhy{}} \PY{n+nc}{proposition}\PY{p}{)}
	      \PY{c+c1}{;; control predicates}
	      \PY{p}{(}\PY{n+nx}{idle}\PY{p}{)}
	      \PY{p}{(}\PY{n+nx}{check\PYZhy{}pre} \PY{n+nv}{?a} \PY{n+nc}{\PYZhy{}} \PY{n+nc}{action} \PY{n+nv}{?p} \PY{n+nc}{\PYZhy{}} \PY{n+nc}{proposition}\PY{p}{)}
	      \PY{p}{(}\PY{n+nx}{apply\PYZhy{}add} \PY{n+nv}{?a} \PY{n+nc}{\PYZhy{}} \PY{n+nc}{action} \PY{n+nv}{?p} \PY{n+nc}{\PYZhy{}} \PY{n+nc}{proposition}\PY{p}{)}
	      \PY{p}{(}\PY{n+nx}{apply\PYZhy{}del} \PY{n+nv}{?a} \PY{n+nc}{\PYZhy{}} \PY{n+nc}{action} \PY{n+nv}{?p} \PY{n+nc}{\PYZhy{}} \PY{n+nc}{proposition}\PY{p}{)}
	      \PY{p}{)}

  \PY{p}{(}\PY{k}{:action} \PY{n+nx}{check\PYZhy{}first\PYZhy{}pre}
   \PY{k}{:parameters} \PY{p}{(}\PY{n+nv}{?a} \PY{n+nc}{\PYZhy{}} \PY{n+nc}{action} \PY{n+nv}{?p} \PY{n+nc}{\PYZhy{}} \PY{n+nc}{proposition}\PY{p}{)}
   \PY{k}{:precondition} \PY{p}{(}\PY{n+nb}{and} \PY{p}{(}\PY{n+nx}{idle}\PY{p}{)} \PY{p}{(}\PY{n+nx}{first\PYZhy{}pre} \PY{n+nv}{?a} \PY{n+nv}{?p}\PY{p}{)} \PY{p}{(}\PY{n+nx}{true} \PY{n+nv}{?p}\PY{p}{)}\PY{p}{)}
   \PY{k}{:effect} \PY{p}{(}\PY{n+nb}{and} \PY{p}{(}\PY{n+nb}{not} \PY{p}{(}\PY{n+nx}{idle}\PY{p}{)}\PY{p}{)} \PY{p}{(}\PY{n+nx}{check\PYZhy{}pre} \PY{n+nv}{?a} \PY{n+nv}{?p}\PY{p}{)}\PY{p}{)}\PY{p}{)}

  \PY{p}{(}\PY{k}{:action} \PY{n+nx}{check\PYZhy{}next\PYZhy{}pre}
   \PY{k}{:parameters} \PY{p}{(}\PY{n+nv}{?a} \PY{n+nc}{\PYZhy{}} \PY{n+nc}{action} \PY{n+nv}{?p} \PY{n+nc}{\PYZhy{}} \PY{n+nc}{proposition} \PY{n+nv}{?q} \PY{n+nc}{\PYZhy{}} \PY{n+nc}{proposition}\PY{p}{)}
   \PY{k}{:precondition} \PY{p}{(}\PY{n+nb}{and} \PY{p}{(}\PY{n+nx}{check\PYZhy{}pre} \PY{n+nv}{?a} \PY{n+nv}{?p}\PY{p}{)} \PY{p}{(}\PY{n+nx}{next\PYZhy{}pre} \PY{n+nv}{?a} \PY{n+nv}{?p} \PY{n+nv}{?q}\PY{p}{)} \PY{p}{(}\PY{n+nx}{true} \PY{n+nv}{?q}\PY{p}{)}\PY{p}{)}
   \PY{k}{:effect} \PY{p}{(}\PY{n+nb}{and} \PY{p}{(}\PY{n+nb}{not} \PY{p}{(}\PY{n+nx}{check\PYZhy{}pre} \PY{n+nv}{?a} \PY{n+nv}{?p}\PY{p}{)}\PY{p}{)} \PY{p}{(}\PY{n+nx}{check\PYZhy{}pre} \PY{n+nv}{?a} \PY{n+nv}{?q}\PY{p}{)}\PY{p}{)}\PY{p}{)}

  \PY{p}{(}\PY{k}{:action} \PY{n+nx}{skip\PYZhy{}check\PYZhy{}pre}
   \PY{k}{:parameters} \PY{p}{(}\PY{n+nv}{?a} \PY{n+nc}{\PYZhy{}} \PY{n+nc}{action} \PY{n+nv}{?p} \PY{n+nc}{\PYZhy{}} \PY{n+nc}{proposition}\PY{p}{)}
   \PY{k}{:precondition} \PY{p}{(}\PY{n+nb}{and} \PY{p}{(}\PY{n+nx}{idle}\PY{p}{)} \PY{p}{(}\PY{n+nx}{has\PYZhy{}no\PYZhy{}pre} \PY{n+nv}{?a}\PY{p}{)} \PY{p}{(}\PY{n+nx}{first\PYZhy{}del} \PY{n+nv}{?a} \PY{n+nv}{?p}\PY{p}{)}\PY{p}{)}
   \PY{k}{:effect} \PY{p}{(}\PY{n+nb}{and} \PY{p}{(}\PY{n+nb}{not} \PY{p}{(}\PY{n+nx}{idle}\PY{p}{)}\PY{p}{)} \PY{p}{(}\PY{n+nx}{apply\PYZhy{}del} \PY{n+nv}{?a} \PY{n+nv}{?p}\PY{p}{)}\PY{p}{)}\PY{p}{)}

  \PY{p}{(}\PY{k}{:action} \PY{n+nx}{apply\PYZhy{}first\PYZhy{}del}
   \PY{k}{:parameters} \PY{p}{(}\PY{n+nv}{?a} \PY{n+nc}{\PYZhy{}} \PY{n+nc}{action} \PY{n+nv}{?p} \PY{n+nc}{\PYZhy{}} \PY{n+nc}{proposition} \PY{n+nv}{?q} \PY{n+nc}{\PYZhy{}} \PY{n+nc}{proposition}\PY{p}{)}
   \PY{k}{:precondition} \PY{p}{(}\PY{n+nb}{and} \PY{p}{(}\PY{n+nx}{check\PYZhy{}pre} \PY{n+nv}{?a} \PY{n+nv}{?p}\PY{p}{)} \PY{p}{(}\PY{n+nx}{last\PYZhy{}pre} \PY{n+nv}{?a} \PY{n+nv}{?p}\PY{p}{)} \PY{p}{(}\PY{n+nx}{first\PYZhy{}del} \PY{n+nv}{?a} \PY{n+nv}{?q}\PY{p}{)}\PY{p}{)}
   \PY{k}{:effect} \PY{p}{(}\PY{n+nb}{and} \PY{p}{(}\PY{n+nb}{not} \PY{p}{(}\PY{n+nx}{check\PYZhy{}pre} \PY{n+nv}{?a} \PY{n+nv}{?p}\PY{p}{)}\PY{p}{)} \PY{p}{(}\PY{n+nx}{apply\PYZhy{}del} \PY{n+nv}{?a} \PY{n+nv}{?q}\PY{p}{)} \PY{p}{(}\PY{n+nb}{not} \PY{p}{(}\PY{n+nx}{true} \PY{n+nv}{?q}\PY{p}{)}\PY{p}{)}\PY{p}{)}\PY{p}{)}

  \PY{p}{(}\PY{k}{:action} \PY{n+nx}{apply\PYZhy{}next\PYZhy{}del}
   \PY{k}{:parameters} \PY{p}{(}\PY{n+nv}{?a} \PY{n+nc}{\PYZhy{}} \PY{n+nc}{action} \PY{n+nv}{?p} \PY{n+nc}{\PYZhy{}} \PY{n+nc}{proposition} \PY{n+nv}{?q} \PY{n+nc}{\PYZhy{}} \PY{n+nc}{proposition}\PY{p}{)}
   \PY{k}{:precondition} \PY{p}{(}\PY{n+nb}{and} \PY{p}{(}\PY{n+nx}{apply\PYZhy{}del} \PY{n+nv}{?a} \PY{n+nv}{?p}\PY{p}{)} \PY{p}{(}\PY{n+nx}{next\PYZhy{}del} \PY{n+nv}{?a} \PY{n+nv}{?p} \PY{n+nv}{?q}\PY{p}{)}\PY{p}{)}
   \PY{k}{:effect} \PY{p}{(}\PY{n+nb}{and} \PY{p}{(}\PY{n+nb}{not} \PY{p}{(}\PY{n+nx}{apply\PYZhy{}del} \PY{n+nv}{?a} \PY{n+nv}{?p}\PY{p}{)}\PY{p}{)} \PY{p}{(}\PY{n+nx}{apply\PYZhy{}del} \PY{n+nv}{?a} \PY{n+nv}{?q}\PY{p}{)} \PY{p}{(}\PY{n+nb}{not} \PY{p}{(}\PY{n+nx}{true} \PY{n+nv}{?q}\PY{p}{)}\PY{p}{)}\PY{p}{)}\PY{p}{)}

  \PY{p}{(}\PY{k}{:action} \PY{n+nx}{skip\PYZhy{}apply\PYZhy{}del}
   \PY{k}{:parameters} \PY{p}{(}\PY{n+nv}{?a} \PY{n+nc}{\PYZhy{}} \PY{n+nc}{action} \PY{n+nv}{?p} \PY{n+nc}{\PYZhy{}} \PY{n+nc}{proposition} \PY{n+nv}{?q} \PY{n+nc}{\PYZhy{}} \PY{n+nc}{proposition}\PY{p}{)}
   \PY{k}{:precondition} \PY{p}{(}\PY{n+nb}{and} \PY{p}{(}\PY{n+nx}{check\PYZhy{}pre} \PY{n+nv}{?a} \PY{n+nv}{?p}\PY{p}{)} \PY{p}{(}\PY{n+nx}{last\PYZhy{}pre} \PY{n+nv}{?a} \PY{n+nv}{?p}\PY{p}{)} \PY{p}{(}\PY{n+nx}{has\PYZhy{}no\PYZhy{}del} \PY{n+nv}{?a}\PY{p}{)} \PY{p}{(}\PY{n+nx}{first\PYZhy{}add} \PY{n+nv}{?a} \PY{n+nv}{?p}\PY{p}{)}\PY{p}{)}
   \PY{k}{:effect} \PY{p}{(}\PY{n+nb}{and} \PY{p}{(}\PY{n+nb}{not} \PY{p}{(}\PY{n+nx}{check\PYZhy{}pre} \PY{n+nv}{?a} \PY{n+nv}{?p}\PY{p}{)}\PY{p}{)} \PY{p}{(}\PY{n+nx}{apply\PYZhy{}add} \PY{n+nv}{?a} \PY{n+nv}{?q}\PY{p}{)}\PY{p}{)}\PY{p}{)}

  \PY{p}{(}\PY{k}{:action} \PY{n+nx}{skip\PYZhy{}check\PYZhy{}pre\PYZhy{}and\PYZhy{}apply\PYZhy{}del}
   \PY{k}{:parameters} \PY{p}{(}\PY{n+nv}{?a} \PY{n+nc}{\PYZhy{}} \PY{n+nc}{action} \PY{n+nv}{?p} \PY{n+nc}{\PYZhy{}} \PY{n+nc}{proposition}\PY{p}{)}
   \PY{k}{:precondition} \PY{p}{(}\PY{n+nb}{and} \PY{p}{(}\PY{n+nx}{idle}\PY{p}{)} \PY{p}{(}\PY{n+nx}{has\PYZhy{}no\PYZhy{}pre} \PY{n+nv}{?a}\PY{p}{)} \PY{p}{(}\PY{n+nx}{has\PYZhy{}no\PYZhy{}del} \PY{n+nv}{?a}\PY{p}{)} \PY{p}{(}\PY{n+nx}{first\PYZhy{}add} \PY{n+nv}{?a} \PY{n+nv}{?p}\PY{p}{)}\PY{p}{)}
   \PY{k}{:effect} \PY{p}{(}\PY{n+nb}{and} \PY{p}{(}\PY{n+nb}{not} \PY{p}{(}\PY{n+nx}{idle}\PY{p}{)}\PY{p}{)} \PY{p}{(}\PY{n+nx}{apply\PYZhy{}add} \PY{n+nv}{?a} \PY{n+nv}{?q}\PY{p}{)}\PY{p}{)}\PY{p}{)}

  \PY{p}{(}\PY{k}{:action} \PY{n+nx}{apply\PYZhy{}first\PYZhy{}add}
   \PY{k}{:parameters} \PY{p}{(}\PY{n+nv}{?a} \PY{n+nc}{\PYZhy{}} \PY{n+nc}{action} \PY{n+nv}{?p} \PY{n+nc}{\PYZhy{}} \PY{n+nc}{proposition} \PY{n+nv}{?q} \PY{n+nc}{\PYZhy{}} \PY{n+nc}{proposition}\PY{p}{)}
   \PY{k}{:precondition} \PY{p}{(}\PY{n+nb}{and} \PY{p}{(}\PY{n+nx}{apply\PYZhy{}del} \PY{n+nv}{?a} \PY{n+nv}{?p}\PY{p}{)} \PY{p}{(}\PY{n+nx}{last\PYZhy{}del} \PY{n+nv}{?a} \PY{n+nv}{?p}\PY{p}{)} \PY{p}{(}\PY{n+nx}{first\PYZhy{}add} \PY{n+nv}{?a} \PY{n+nv}{?q}\PY{p}{)}\PY{p}{)}
   \PY{k}{:effect} \PY{p}{(}\PY{n+nb}{and} \PY{p}{(}\PY{n+nb}{not} \PY{p}{(}\PY{n+nx}{apply\PYZhy{}del} \PY{n+nv}{?a} \PY{n+nv}{?p}\PY{p}{)}\PY{p}{)} \PY{p}{(}\PY{n+nx}{apply\PYZhy{}add} \PY{n+nv}{?a} \PY{n+nv}{?q}\PY{p}{)} \PY{p}{(}\PY{n+nx}{true} \PY{n+nv}{?q}\PY{p}{)}\PY{p}{)}\PY{p}{)}

  \PY{p}{(}\PY{k}{:action} \PY{n+nx}{apply\PYZhy{}next\PYZhy{}add}
   \PY{k}{:parameters} \PY{p}{(}\PY{n+nv}{?a} \PY{n+nc}{\PYZhy{}} \PY{n+nc}{action} \PY{n+nv}{?p} \PY{n+nc}{\PYZhy{}} \PY{n+nc}{proposition} \PY{n+nv}{?q} \PY{n+nc}{\PYZhy{}} \PY{n+nc}{proposition}\PY{p}{)}
   \PY{k}{:precondition} \PY{p}{(}\PY{n+nb}{and} \PY{p}{(}\PY{n+nx}{apply\PYZhy{}add} \PY{n+nv}{?a} \PY{n+nv}{?p}\PY{p}{)} \PY{p}{(}\PY{n+nx}{next\PYZhy{}add} \PY{n+nv}{?a} \PY{n+nv}{?p} \PY{n+nv}{?q}\PY{p}{)}\PY{p}{)}
   \PY{k}{:effect} \PY{p}{(}\PY{n+nb}{and} \PY{p}{(}\PY{n+nb}{not} \PY{p}{(}\PY{n+nx}{apply\PYZhy{}add} \PY{n+nv}{?a} \PY{n+nv}{?p}\PY{p}{)}\PY{p}{)} \PY{p}{(}\PY{n+nx}{apply\PYZhy{}add} \PY{n+nv}{?a} \PY{n+nv}{?q}\PY{p}{)} \PY{p}{(}\PY{n+nx}{true} \PY{n+nv}{?q}\PY{p}{)}\PY{p}{)}\PY{p}{)}

  \PY{p}{(}\PY{k}{:action} \PY{n+nx}{finish}
   \PY{k}{:parameters} \PY{p}{(}\PY{n+nv}{?a} \PY{n+nc}{\PYZhy{}} \PY{n+nc}{action} \PY{n+nv}{?p} \PY{n+nc}{\PYZhy{}} \PY{n+nc}{proposition}\PY{p}{)}
   \PY{k}{:precondition} \PY{p}{(}\PY{n+nb}{and} \PY{p}{(}\PY{n+nx}{apply\PYZhy{}add} \PY{n+nv}{?a} \PY{n+nv}{?p}\PY{p}{)} \PY{p}{(}\PY{n+nx}{last\PYZhy{}add} \PY{n+nv}{?a} \PY{n+nv}{?p}\PY{p}{)}\PY{p}{)}
   \PY{k}{:effect} \PY{p}{(}\PY{n+nb}{and} \PY{p}{(}\PY{n+nb}{not} \PY{p}{(}\PY{n+nx}{apply\PYZhy{}add} \PY{n+nv}{?a} \PY{n+nv}{?p}\PY{p}{)}\PY{p}{)} \PY{p}{(}\PY{n+nx}{idle}\PY{p}{)}\PY{p}{)}\PY{p}{)}
  \PY{p}{)}
\end{Verbatim}
\caption{A STRIPS formulation of the universal propositional planning domain.}
\label{fig:upp:strips}
\end{figure}

A STRIPS formulation of the universal propositional planning domain,
without quantified or conditional effects, can be obtained using the
idea from Nebel's \citeyear{nebel00} polynomial size compilation, but
simplified because all predicates in effect conditions are static.
A possible formulation is shown in Figure \ref{fig:upp:strips}.
This formulation assumes each action has at least one add effect.
In this formulation, application of a ground action $a$ is done by
first sequentially checking each of its precondition propositions hold
in the current state, then sequentially applying its delete and add
effects. The predicate \ptype{idle} represents that no action application
is in progress, and must be true in the initial state and goal of an
instance of this domain.

\begin{figure}
\small
\begin{Verbatim}[commandchars=\\\{\}]
\PY{p}{(}\PY{k}{define} \PY{p}{(}\PY{k}{domain} \PY{n+nx}{parameterised-strips-planning-3-2-1}\PY{p}{)}
  \PY{p}{(}\PY{k}{:predicates} \PY{p}{(}\PY{n+nx}{ground-action} \PY{n+nv}{?pre1} \PY{n+nv}{?pre2} \PY{n+nv}{?pre3} \PY{n+nv}{?add1} \PY{n+nv}{?add2} \PY{n+nv}{?del1}\PY{p}{)}
	       \PY{p}{(}\PY{n+nx}{true} \PY{n+nv}{?p}\PY{p}{)}\PY{p}{)}
  \PY{p}{(}\PY{k}{:action} \PY{n+nx}{apply}
   \PY{k}{:parameters} \PY{p}{(}\PY{n+nv}{?pre1} \PY{n+nv}{?pre2} \PY{n+nv}{?pre3} \PY{n+nv}{?add1} \PY{n+nv}{?add2} \PY{n+nv}{?del1}\PY{p}{)}
   \PY{k}{:precondition} \PY{p}{(}\PY{n+nb}{and} \PY{p}{(}\PY{n+nx}{ground-action} \PY{n+nv}{?pre1} \PY{n+nv}{?pre2} \PY{n+nv}{?pre3} \PY{n+nv}{?add1} \PY{n+nv}{?add2} \PY{n+nv}{?del1}\PY{p}{)}
                      \PY{p}{(}\PY{n+nx}{true} \PY{n+nv}{?pre1}\PY{p}{)} \PY{p}{(}\PY{n+nx}{true} \PY{n+nv}{?pre2}\PY{p}{)} \PY{p}{(}\PY{n+nx}{true} \PY{n+nv}{?pre3}\PY{p}{)}\PY{p}{)}
   \PY{k}{:effect} \PY{p}{(}\PY{n+nb}{and} \PY{p}{(}\PY{n+nx}{true} \PY{n+nv}{?add1}\PY{p}{)} \PY{p}{(}\PY{n+nx}{true} \PY{n+nv}{?add2}\PY{p}{)} \PY{p}{(}\PY{n+nb}{not} \PY{p}{(}\PY{n+nx}{true} \PY{n+nv}{?del1}\PY{p}{)}\PY{p}{)}\PY{p}{)}
   \PY{p}{)}
  \PY{p}{)}
\end{Verbatim}
\caption{A STRIPS formulation of the parameterised universal planning domain $D_{3,2,1}$.}
\label{fig:parameterised}
\end{figure}

Alternatively, we can construct a \emph{parameterised} universal PDDL domain, $D_{p,a,d}$, with one single action schema, two predicate symbols, and no types, using only the STRIPS subset of PDDL.
Instances of this domain can encode all propositional planning tasks with at most $p$ propositions in any actions' precondition, $a$ propositions in any add effect, and $d$ propositions in any delete effect.
Figure~\ref{fig:parameterised} shows the domain $D_{3,2,1}$ (i.e., parameterised universal domain with $p=3, a=2, d=1$).
In an instance $P^{D_{3,2,1}}_{\Pi}$ of this universal domain, the set of objects corresponds to the propositions of $\Pi$. Each initial state atom with predicate symbol \ptype{ground-action} exactly describes a ground
action of $\Pi$. Moreover, the single action schema $\mptype{apply}$ checks that the preconditions of the ground action are true in the current state, and applies the effects according to the ground action.

Note that our example $D_{3,2,1}$ is already sufficient to represent common
encodings of Turing Machines as propositional planning problems
\cite{bylander94}. In fact, even $D_{2,1,1}$ is sufficient.
Therefore, we can immediately conclude a few complexity results for lifted planning, for example, that it is enough to have one single action schema and one fluent (i.e., non-static) predicate to reach PSPACE-hardness.

\section{The Impossibility of a Lifted Universal PDDL Domain}

One of the few restrictions that a PDDL domain does impose on all instances
of the domain is a fixed maximum arity of predicates. This implies a limit
on the length of the shortest plans required to solve any instance of the domain.
Let $D$ be a PDDL domain description, and suppose the maximum arity of any
predicate in $D$ is $k$. Let $P$ be a PDDL problem description, that is an
instance of $D$, and $m$ the number of objects in $P$.
Note that $k \leq |D|$ and $m \leq |P|$.
We know that the grounding of $(D,P)$ can have at most $m^k$ propositions,
and therefore the length of a shortest plan for $(D,P)$ is bounded by
$2^{(m^k)}$.

However, we also know that it is possible to construct a PDDL domain $D'$
and problem $P'$ requiring a shortest plan of length $2^{2^{(n-1)}}$, where
$|D|+|P| \leq c (n \log n)$, for some constant $c$.
(This construction was first demonstrated by \citeA{erol-nau-subrahmanian91},
and can also be found in Section 2.5.4 of the book by
\citeA{haslum-etal:pddl19}.)

Suppose there exists a universal domain for lifted PDDL planning:
the maximum arity of any predicate in this domain is a fixed constant $k$.
Therefore, an instance $P_{(D',P')}$ of this domain representing the domain
and problem $(D',P')$ must have a number of objects $m$ such that
${m^k} \geq 2^{n-1}$, implying $m \geq 2^{\nicefrac{(n-1)}{k}}$, and hence that the
size of $P_{(D',P')}$ must grow exponentially in the size of $D'$ and $P'$.

Note, however, that a more expressive formalism than classical PDDL, such
as, for example, a language with recursive terms (function symbols), may
not have a corresponding bound on plan length, and therefore may well have
the capacity to formulate a universal domain that admits any domain--problem
pair as a polynomial-size instance.

\section{Discussion}

What are the implications of the existence of the universal planning domain?

First, it demonstrates that domain-dependent planning, or generalised planning,
is, in the general case, and with PDDL's definition of domain, futile: there
exists domains for which there is no domain-specific strategy or solution
algorithm more efficient than one that works for every domain, and for which
the only generalised plan that exists is a domain-independent planner.

We can of course classify domains by the computational complexity (hardness) of generalised or domain-specific planning for them, i.e., of the set of instances that they admit. Such studies have been made on a range of commonly used planning benchmark domains \cite{helmert01,helmert:icaps-2006}.
A very small number of works have identified fragments of lifted PDDL planning that are tractable. For example, \citeA{lauer-etal:IJCAI:2021} show that delete-free planning in a domain with at-most-unary predicates is tractable.
Our universal domain constructions complement these results, by showing that domains remain PSPACE-hard also under a variety of syntactic restrictions.
Tractability of a restricted class of domains, however, is typically obtained from the existence of a general, i.e., domain-independent, polynomial-time algorithm that solves instances of domains satisfying the restriction.

The impact on the complexity of domain-specific lifted or generalised planning of restrictions in between these two, somewhat extreme, cases -- analysing specific domains vs.\ coarse syntactic restrictions on PDDL domain formulations -- is, as far as we know, mostly unexplored.
\citeA{jonsson-backstrom:aaai:1996} make an important observation: They study the existence of a ``universal plan'' (essentially, policy) for propositional planning problems, and show that a universal plan that is both compact (polynomial-size) and efficient (evaluable in polynomial time) does not exist for arbitrary propositional planning problems, but can be constructed ``for planning problems such that [optimal-length   plan generation] can be solved in polynomial time''.
If we understand ``planning problems'' here as the family of instances of a given (PDDL) domain, this suggests that generalised planning is possible for, and only for, domains that encode tractable underlying problems. Characterising the class of such domains by syntactic restrictions is likely to be challenging (though perhaps possible via descriptive complexity theory).

\bibliographystyle{theapa}
\bibliography{upd}

\end{document}